\documentclass[journal]{IEEEtran}
\usepackage{algorithm}
\usepackage{algorithmic}
\usepackage{graphicx}
\usepackage{caption,subcaption}
\usepackage{float}
\usepackage{amsmath}
\usepackage{amssymb}
\usepackage{mathrsfs}
\usepackage{booktabs}
\usepackage{bbm}

\usepackage[pagebackref,breaklinks,colorlinks]{hyperref}

\ifCLASSINFOpdf
\else
\fi
\begin{document}
%
\title{A Coarse-to-Fine Human Pose Estimation Method based on Two-stage Distillation and Progressive Graph Neural Network}
%
%
%

\author{Zhangjian ~Ji,~\IEEEmembership{Member,~IEEE,}
        Wenjin ~Zhang, Shaotong ~Qiao, Kai ~Feng and Yuhua ~Qian,~\IEEEmembership{Member,~IEEE,}
\thanks{Zhangjian Ji, Wenjin Zhang, Shaotong Qiao and Kai Feng are with the Key Laboratory of Computational Intelligence
and Chinese Information Processing of Ministry of Education, School of Computer and Information Technology, Shanxi University, Taiyuan 030006, China E-mail: \{jizhangjian@sxu.edu.cn\}}
\thanks{Yuhua Qian is with the Institute of Big Data Science
and Industry, the Key Laboratory of Computational Intelligence and
Chinese Information Processing of Ministry of Education, Shanxi University,
Taiyuan 030006, Shanxi, China E-mail: \{ jinchengqyh@sxu.edu.cn \}}
}

%
%

\markboth{Journal of \LaTeX\ Class Files,~Vol.~14, No.~8, November~2022}%
{Shell \MakeLowercase{\textit{et al.}}: Bare Demo of IEEEtran.cls for IEEE Journals}
%



\maketitle

\begin{abstract}
Human pose estimation has been widely applied in the human-centric understanding and generation, but most existing state-of-the-art human pose estimation methods require heavy computational resources for accurate predictions. In order to obtain an accurate, robust yet lightweight human pose estimator, one feasible way is to transfer pose knowledge from a powerful teacher model to a less-parameterized student model by knowledge distillation. However, the traditional knowledge distillation framework does not fully explore the contextual information among human joints. Thus, in this paper, we propose a novel coarse-to-fine two-stage knowledge distillation framework for human pose estimation. In the first-stage distillation, we introduce the human joints structure loss to mine the structural information among human joints so as to transfer high-level semantic knowledge from the teacher model to the student model. In the second-stage distillation, we utilize an Image-Guided Progressive Graph Convolutional Network (IGP-GCN) to refine the initial human pose obtained from the first-stage distillation and supervise the training of the IGP-GCN in the progressive way by the final output pose of teacher model. The extensive experiments on the benchmark dataset: COCO keypoint and CrowdPose datasets, show that our proposed method performs favorably against lots of the existing state-of-the-art human pose estimation methods, especially for the more complex CrowdPose dataset, the performance improvement of our model is more significant.
\end{abstract}


%
\IEEEpeerreviewmaketitle

\section{Introduction}
%
%
%
%
\IEEEPARstart{H}{uman} pose estimation (HPE) has been being a fundamental yet challenging research topic in the computer vision field, which aims at localizing body joints of each person from a single image and describes the human skeleton information by those body joints. It plays an important role in lots of human-centric perception and understanding tasks, including human action recognition \cite{author2}, human-computer interaction \cite{author1}, pose-conditioned human motion generation \cite{author3}, etc. Although human pose estimation has made great progress by advanced deep learning technologies in the past decade, there is still some serious challenges (e.g., illumination variation, object occlusion, self-occlusion, etc.) to design a robust and accurate human pose estimation algorithm in crowd scenarios.

At present, in order to predict human joints accurately, most existing human pose estimation algorithms fucus on learning richer features representation by the various deep network architectures, e.g, HRNet \cite{author26}, HigherHRNet \cite{author25}, ViTPose \cite{author52} and so on. However, the performance gains of these approaches often come at the expense of training and deploying over-parameterized network models, which severely hinders them from applying to resource-constrained mobile devices. Thus, some researchers devoted themselves to design lightweight and real-time network models, \emph{e.g.}, Lite-HRNet \cite{author30}, Dite-HRNet \cite{author31}, MultiPoseNet \cite{author71} and so on. Unfortunately, the performance of lightweight models was also worse relatively.

The knowledge distillation is regarded as an effective way to transfer knowledge from big model to lightweight small model, which enables a small student model to learn better by inheriting knowledge from a large teacher model and has been widely applied in various visual tasks, such as classification \cite{author72}, detection \cite{author73}, and segmentation \cite{author74}. Recently, it has also been successfully used in the human pose estimation task, e.g., DistilPose \cite{author48}, OKDHP \cite{author46}, DWPose \cite{author56}, DOPE \cite{author50}, ViTPose \cite{author52} and so on. Although these methods balanced speed and accuracy better, they didn't consider the inherent topology structure information among human joints in the distillation process. In fact, the structure information among human joints can help locate the human joints more accurately, especially in crowd scenarios which exist some invisible human joints because of occlusion problem. Currently, there has been mined the structure information of human joints to improve their predicting accuracy by graph neural networks \cite{author64} or adding the explicit joints structure constraint in the loss \cite{author49}, but the structure information of human joints has been rarely studied in the distillation models for human pose estimation. Thus, in this paper, we propose a novel two-stage knowledge distillation framework for human pose estimation, which adopts the latest pose estimator SimCC \cite{author70} as the basic model because it can achieve more competitive accuracy with lower computational effort by formulating keypoints prediction as classification from sub-pixel bins for horizontal and vertical coordinates respectively.

In the first-stage distillation, we natively utilized the intermediate layer and final output of the teacher model to supervise the training of student model. Different from the previous distillation models (e.g., DWPose \cite{author56}) that adopted the logit distillation usually used in classification task in the final output, in order to predict the location of invisible human joints in crowd scenarios, we introduced the human joints structure loss and $\ell_{1}$ loss in the distillation of final output to mine the structure information among human joints. One insight is that the invisible joints are strongly related to not only structural understanding of the human pose but also the contextual understanding of the image. For example, human can easily infer the location of those invisible joints using the clues derived from the image context. Therefore, in the second-stage distillation, so as to mine the contextual information among human joints from the image, we introduced an Image-Guided Progressive Graph Convolutional Network (IGP-GCN) \cite{author68} to adjust the human pose obtained from the first-stage distillation. In addition, to improve the accuracy of human joints further, we supervised the training of the IGP-GCN module in the progressive way by the final output of teacher model.

The main contributions of this work are summarized as follows:
\begin{itemize}
  \item We propose a novel two-stage knowledge distillation framework for human pose estimation, which adopts the SimCC that can obtain the sub-pixel accuracy for human joints prediction as the basic model.
  \item We introduce the human joints structure loss and $\ell_{1}$ loss in the distillation of final output to mine the structure information among human joints.
  \item We utilize an IGP-GCN module to adjust the human pose obtained from the first-stage distillation and supervise the training of the IGP-GCN module in the progressive way by the final output pose of teacher model.
  \item Quantitative experiments and ablation study on the authoritative COCO keypoint and CrowdPose datasets show that the proposed model and its each components are reasonable and effective.
\end{itemize}

 The remaining of this work is organized as follows: Section \ref{sec:rw} introduces the related works. In section \ref{sec:PM}, we elaborate the proposed human anatomical keypoints constraint model and how to plug it into the existing bottom-up and top-down methods. Section \ref{sec:exp} shows the ablation study and evaluation results on COCO keypoints datasets. Finally, conclusions are explained in Section \ref{sec:con}.
\section{Related works}\label{sec:rw}
\subsection{Heatmap-based methods for pose estimation}
Generally, the heatmap-based approaches for HPE can be divided into two categories, i.e., bottom-up and top-down methods. The bottom-up methods first predict identity-free keypoints of all the persons from the input image by human anatomical keypoints detection network, and then assign them to the corresponding person intances. The existing bottom-up methods mainly attempted to construct the more efficient network architecture (e.g., HRNet \cite{author26} and HigherHRNet \cite{author25}) and design different keypoint grouping strategies \cite{author21, author22}. In order to localize body joints from a single image and group them into the corresponding persons, Cao \emph{et al. }\cite{author22} adopted a two-branch multi-stage network to predict the human joints heatmaps and the 2D vector field of human keypoint affinities. Newell \emph{et al.}\cite{author23} utilized a stacked hourglass network\cite{author24} to simultaneously produce a detection heatmap and a tagging heatmap for each body joint and group body joints with similar tags into individual people. Li \emph{et al.} \cite{author42} combined the bottom-up pipeline with human detection, which avoided mistake propagation across the poses of different persons. Deepcut \cite{author40} assigned labeling candidate joints to individual persons by integer linear programming and Zanfir \emph{et al.} \cite{author41} grouped body joints by learned scoring functions. Although these bottom-up methods have good performance for HPE, they are quite large and computationally expensive and not suitable for Internet-of-Things (IoT) applications requiring real time and lightweight, and exist a fatal flaw that the invisible body joints will degrade the performance drastically.

In the second category, the top-down methods firstly use an object detector to locate the bounding boxes of all the persons from the input image, and then perform pose estimation on the cropping bounding box region of each person. So most of the existing top-down methods focused on proposing a more effective human detector to obtain better person detection \cite{author11,author43} or how to better carry out single-person pose estimation on the cropping patch of each person \cite{author28,author29}. For example, Mask R-CNN \cite{author11} directly addedd a mask branch on the FasterRCNN \cite{author1} to predict $K$ mask, one for each of $K$ human keypoint types (e.g., left shoulder, right elbow). Fang \emph{et al.} \cite{author43} proposed a novel regional multi-person pose estimation framework to facilitate pose estimation in the presence of inaccurate or redundant human bounding boxes. Papandreou \emph{et al.} \cite{author27} presented a simple, yet powerful, top-down human pose estimation approach consisting of two stages, which first predicted the bounding box of each person from the input image by the FasterRCNN \cite{author44} and then estimated the joints of the person potentially contained in each proposed bounding box. Chen \emph{et al.} \cite{author8} proposed a cascaded pyramid network, which integrated global pyramid network (GlobalNet) and pyramid refined network based on online hard keypoints mining (RefineNet) that were used to predict the simple and hard human joints, respectively. Bin \emph{et al.} \cite{author29} provided a simplest baseline method for the top-down human pose estimation, which added a few deconvolutional layers over the last convolution stage of the ResNet, generating the high-resolution heatmaps for predicting human joints more precisely. In order to handle the association problem of person joints in crowd scenarios, Li \emph{et al.} \cite{author45} builded a person-joint connection graph after detecting the candidate person joints and solved it by a global maximum joints association algorithm. Nevertheless, all these top-down methods depend on the quality of the detected human bounding boxes, because the single person pose estimator is usually sensitive to these ones, especially in the severe occlusion cases, where one bounding box may capture joints of multiple persons or contain partially invisible joints that lack visual information. Furthermore, it is very time-consuming that the runtime of the top-down approaches is proportional to the number of the detected persons.
\subsection{Knowledge distillation methods for pose estimation}
Knowledge distillation, originally proposed by Hinton \emph{et al.} \cite{author54}, is a way to compress model, which aims to transfer knowledge from a large and computational expensive teacher model to a small and computational efficient student model by supervised learning using the soft labels. At present, it has been widely applied to many computer task, including object detection \cite{author57,author58,author59}, semantic segment \cite{author60,author61}, image generation \cite{author55}, human pose estimation \cite{author14,author46,author50,author51,author53,author56} and so on. For 2D human pose estimation, Zhang \emph{et al.} \cite{author51} first adopted the classical knowledge distillation to supervise a lightweight 4-Stack hourglass network with the soft labels of an 8-Stack hourglass network. Li \emph{et al.} \cite{author46} proposed an online pose distillation approach to distill the structure knowledge of human pose in a one-stage manner in order to guarantee the distillation efficiency. DOPE \cite{author50} presented to distill three separate body-part expects' knowledge into a single whole-body pose estimation network. DWPose \cite{author56}, proposed by Yang \emph{et al.}, was a two-stage pose distillation model for 2D whole-body pose estimation. ViTPose \cite{author52} also implemented a large-to-small model knowledge distillation to prove its knowledge transferability. Ye \emph{et al.} \cite{author48} provided a heatmap-to-regression distillation framework, which is the first work to transfer knowledge between heatmap-based and regression-based models losslessly. The distillation methods above need an extra manipulation to align the features vector during feature distillation, which will lead to a potential performance decrease. Based on this, Chen \emph{et al.} \cite{author47} designed a self-distillation paradigm SDPose that extracts the knowledge in the Multi-Cycled Transformer (MCT) module into one single pass model, which balances performance and resource consumption better.
\subsection{Graph neural network methods for pose estimation}
Graph neural networks (GNN) were originally designed to process the graph analysis tasks such as node classification \cite{author62} and link prediction \cite{author63}. The human joints exhibit a natural graph structure, so some advanced human pose estimation methods \cite{author64,author65,author66} constructed graph neural networks to capture dependency relationships between human joints. For example, in order to estimate the human pose better, Qiu et al. \cite{author65} modeled the relations of different joints to enhance their feature. Bin \emph{et al.} \cite{author64} explored the power of graph convolutional networks to explicitly model the structured relationships between human joints. PINet \cite{author66} introduced GNN to estimate the human joints under the occlusion, which can rectify the inaccurate human joints to a certain extent. Zheng \emph{et al.} \cite{author67} proposed a hierarchical graph neural network for human hose estimation, which can learn the correlation between keypoints explicitly and use the learned correlation for human pose estimation. OPEC-Net \cite{author68} estimated the invisible joints from an inference perspective by proposing an Image-Guided Progressive GNN module which provided a comprehensive understanding of both image context and pose structure.
\subsection{Other methods}
In addition to several human pose estimation methods mentioned above, there are other methods to directly regress the location of human joints from the input image. Although the overall performance of such methods is usually worse, they can truly achieve the end-to-end training and a few can also reach the accuracy of the top-down human pose estimation methods. For example, the CenterNet \cite{author5} directly regressed the location of human anatomical keypoints from the human center and the disentangled keypoint regression (DEKR) \cite{author7} proposed a separate multi-branch regression scheme, each branch of which independently learned a representation with dedicated adaptive convolutions for each keypoint and regressed the position of the corresponding keypoint. SimCC \cite{author70} presented a simple yet effective coordinate classification pipeline for human pose estimation, which regarded human joints prediction as two classification task from sub-pixel bins for horizontal and vertical coordinates respectively. Similar to anchor-based object detection, the PointSetNet \cite{author6} adopted a series of point-set anchors which follow the pose distribution of the training data to initialize the human pose in the input image and predicted the offsets of point-set anchors for real human joints in the training process. As a good task-specific initialization, the point set can yield features that better facilitate human joints localization.
\section{Method}\label{sec:PM}
\begin{figure*}
  \centering
  \includegraphics[width=16cm]{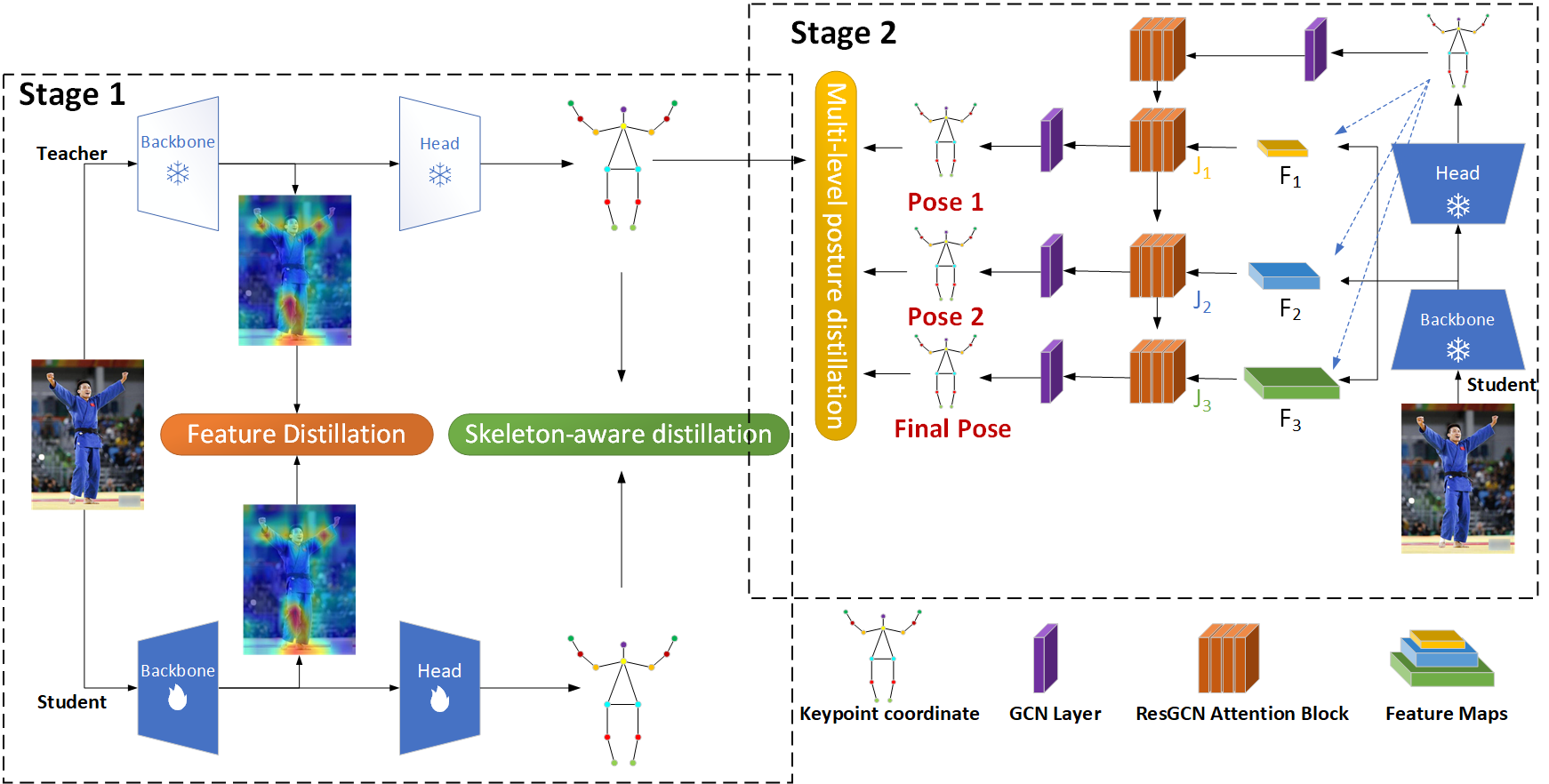}\\
  \caption{The overall work-flow of our proposed two-stage distillation human pose estimation method.}\label{fig_1}
\end{figure*}

In this section, as shown in Figure \ref{fig_1}, we specifically describe our proposed coarse-to-fine human pose estimation method, which combines two-stage knowledge distillation and progressive graph neural network. In the first-stage distillation, we utilize the output more rich feature and more accurate pose of a pre-trained teacher model to guide the student model from scratch. In the second stage, in order to refine the result of human pose estimation further, we freeze the first-stage trained student model, and its output features and pose are feeded into an IGP-GCN module that is supervised in the progressive way by the final output pose of teacher model.
\subsection{The first-stage distillation}\label{sec:l1cft}
In this stage, we force the student model to learn knowledge from the teacher model by the feature-based and pose structure-based distillations, which can utilize the rich information of the teacher model to enhance the student's performance.
\subsubsection{Feature-based distillation}
The feature distillation aims to force the output features of the student's backbone to mimic the output ones of the teacher's directly, so that the student network can learn richer feature representations. In the distillation processing, considering that the number of feature channels output from the student's and the teacher's backbones may be inconsistent, we first utilize the $1\times1$ convolution to align them. Then, to transfer knowledge from the output features of the teacher's backbone, we use the  mean square error (MSE) loss to calculate the distance between the student's feature $F^{stu}$and the teacher's feature $F^{tea}$, which can be formulated as:
\begin{equation}\label{eq1}
  L_{fea}=\frac{1}{CHW}\sum_{c=1}^{C}\sum_{h=1}^{H}\sum_{w=1}^{W}\left(F_{c,h,w}^{tea}-{cov_{1\times1}(F^{stu})}_{c,h,w}\right)^{2},
\end{equation}
where $cov_{1\times1}$ describes the $1\times1$ convolution operation, and it reshapes the $F^{stu}$ to the same dimension as the $F^{tea}$. $H,W,C$ respectively denote the width, height and channel of the features output from the teacher's backbone.

\subsubsection{Pose structure distillation}
The pose structure distillation aims to make the human keypoints predicted by the student network as close as possible to the teacher's. We design the following human pose distillation loss function, which can be optimized so that the student network can learn more accurate predicting ability for human joints from the teacher one.
\begin{equation}\label{eq-pos}
 L_{pos}=\frac{1}{N}\sum_{n=1}^{N}|P_{n}^{tea}-P_{n}^{stu}|+L_{cst},
\end{equation}
where $N$ denotes the number of the training samples, $P_{n}^{tea}$ and $P_{n}^{stu}$ respectively represent the sets consisting of each human joint location predicted by the teacher and student models, and the second term $L_{cst}$ of the Eq.(\ref{eq-pos}) is the structural constraint loss of human pose, which is a simplified version of the human anatomical keypoints constraint model proposed in \cite{author49}, and can be defined as
\begin{equation}\label{eq-cst}
L_{cst}=\sum_{(i,j)\in \mathcal{E}}\omega_{i,j}|\mathbf{\hat{e}}_{i,j}^{tea}-\mathbf{\hat{e}}_{i,j}^{stu}|,
\end{equation}
where $\mathcal{E}$ denotes the connecting edge set of human keypoints that have the connecting relation, $\mathbf{\hat{e}}_{i,j}$ denotes the length of the connecting edge between the $i$th and $j$th human keypoints (if the $i$th and $j$th krypoints are connected), and $\omega_{i,j}$ is the weight that measures the contribution of the different edges (Its definition is the same as that of the corresponding weight in \cite{author49}).
\subsubsection{Weight-decay strategy for distillation}
The knowledge distillation aims to transfer knowledge from the teacher network to the student one. However,  the over-fitting can result in the student network that just mimics by rote the teacher's output and fails to generalize to the new data. Inspirited by a detection distillation method TADF \cite{author75}, during the first-stage distillation, we use the weight-decay strategy to reduce the distillation penalty gradually, which can improve the generalization of the student network. The core idea of weight-decay strategy is to introduce a time function $\gamma(t)$, which makes the weight of the distillation loss decrease gradually with the training processing. Specifically, the time function $\gamma(t)$ is as follows:
\begin{equation}\label{eq2}
 \gamma(t)=1-(t-1)/T,
\end{equation}
where $t\in(1,\cdots,T)$ is the current epoch and $T$ denotes the total epochs for training.

By this way, the distillation losses dominate in the early stage of training, which can transfer knowledge from the teacher model more efficiently. As the training goes on, the weights of the distillation losses are gradually decreased, which makes the student network to focus on itself predictive ability so as to improve its generalization. In the coarse-grained training phase, besides the feature distillation and pose structure distillation losses, we select the Kullback-Leibler divergence loss that adopted in the SimCC as the original loss of the student model. Thus, based on the weight-decay strategy, the total losses of the first-stage training are as follows:
\begin{equation}\label{eq3}
 L_{s1}=L_{KL}+\gamma(t)\cdot\alpha L_{fea}+\gamma(t)\cdot\beta L_{pos},
\end{equation}
where $L_{KL}$ denotes the Kullback-Leibler divergence loss and $\alpha, \beta$ are hyper-parameters to balance the loss.
\subsection{The second-stage fine-grained pose distillation}
In order to refine the performance of human joints prediction further, inspirited by the OPEC \cite{author68},  in the seconde distillation stage, we introduce an IGP-GCN module to adjust the human pose obtained from the first-stage distillation and supervise the training of the IGP-GCN module in the progressive way by the final output pose of teacher model.
\subsubsection{Image-guided progressive graph convolution network}
The human body skeleton has a natural hierarchical graph structure, so the relationship between different joints should be helpful to predict the position of human joints more accurately. Based on this,  we construct an undirected graph $G=(V,E)$ to formulate the human pose with $K$ joints, where $V=\{v_{i}\mid i=1,2,\cdots,K\}$ and $E=\{v_{i}v_{j}\mid $ if $i$ and $j$ are connected in the human body\} respectively denote the node set and edge set (the limbs of human body) in $G$. The adjacent matrix of $G$ is $A=\{a_{i,j}\}$, where $a_{i,j}=1$ when $v_{i}$ and $v_{j}$ have the connected edge in $G$, otherwise, $a_{i,j}=0$.

Because the invisible joints lack the visual information, the estimated position of invisible joints from the base module is sometimes far from their correct locations. To refine the position of invisible joints, we design an intuitive coarse-to-fine learning mechanism in the coordinate-based module, which builds an image-guided progressive graph convolution network (IGP-GCN) architecture. As shown in the Figure \ref{fig_1}, this architecture first inputs the initial pose structure clues $P_{init}$ of human joints that estimated from the first-stage trained student model into the first graph convolution layer of the IGP-GCN, and then feed the three node feature vectors $\hat{J}_{i},i=1,2,3$ that consisted of the feature of each human joint excavated from the three low-to high resolution feature maps $\hat{F}_{i},i=1,2,3$ into the second-to-fourth residual graph convolution network (resGCN) attention blocks from coarse to fine orders.
\subsubsection{Fine pose distillation}
Furthermore, in order to refine the estimated human pose, we supervise the training of IGP-GCN module in the progressive manner by the final output pose of the teacher model. During training, we freeze the trained student model from the first-stage distillation. Then, the output pose $P_{i}$ of the trained student model and the three node feature vectors $\hat{J}_{i},i=1,2,3$ are fed into the IGP-GCN module accordingly, where the $\hat{J}_{i},i=1,2,3$ consist of the features located in each joint coordination $<x_{i}^{j},y_{i}^{j}>,j=1,2,\cdots,K$ of human pose $P_{i}$ on the related three low-hight resolution feature maps $\hat{F}_{i},i=1,2,3$ which are output from the trained student's backbone. Finally, we can obtain the predicted human pose $\hat{P}_{i}^{stu}$ under three different resolution feature maps from the IGP-GCN module. For refining the human pose by the second-stage distillation, we design the following loss function, that is
\begin{equation}\label{eq4}
 L_{i}^{GCN}=\frac{1}{N}\sum_{n=1}^{N}|(P_{n}^{tea}-\hat{P}_{i,n}^{stu})\cdot M|,
\end{equation}
where $\hat{P}_{i,n}^{stu}, i\in\{1,2,3\}$ respectively denotes the output human pose from the second-fourth resGCN attention blocks of the IGP-GCN module for the $n$th training sample, $N$ is the total number of the training samples, $P_{n}^{tea}$ represents the output human pose of the teacher model for the $n$th training sample. $M\in\mathbb{Z}_{2}^{K}$ is a binary mask where the element in $M$ corresponds to 1 when the related joint has a ground truth label, otherwise, it is 0. The $\cdot$ denotes the element-wise product operation so that we only take into account the errors on the joints with ground truth. Furthermore, by weighing the distillation loss $L_{_i}^{GCN}$ corresponding to three different scales feature maps, the final loss of the second-stage distillation can be formulated as
\begin{equation}\label{eq5}
  L_{s2}=\delta L_{1}^{GCN}+\lambda L_{2}^{GCN} +\xi L_{3}^{GCN},
\end{equation}
where $\delta$, $\lambda$ and $\xi$ are the hyper-parameters for balancing three losses.
\section{Experiments}\label{sec:exp}
In this section, we first introduce the benchmark datasets and evaluation metrics used in our experiments. Next, we describe the implementation details. Then, we conduct ablation study to analyze the effect of the two-stage distillation and the IGP-GCN module in our proposed human joint estimation method. Finally, we also compare our proposed model with other some state-of-the-art and the most related human pose estimation methods on the MS COCO and CrowdPose datasets.
\subsection{Experimental setup}
\noindent\textbf{Datasets.} The MS COCO keypoint detection dataset (in 2017) \cite{author36} is one of the largest scale and most challenging datasets for human pose estimation, which contains over 200K images and 150K person instances labels with 17 keypoints and is divided into \emph{train/val/test-dev} sets respectively. In addition, for evaluating the performance of our proposed method in dense pose scenes, we also conduct comparative experiments in the CrowdPose dataset, which contains much more crowded scenes than the COCO keypoint dataset. There are 20K images and 80K person instance in CrowdPose dataset, which is split into training, validation and testing subset consist of about 10K, 2K, and 8K images respectively. We report the results on the CrowdPose-test set for ablation studies and also give the comparison results on the COCO \emph{val2017}, COCO \emph{test-dev2017} and CrowdPose-test sets. For training our proposed model, in this paper, we follow the data augmentation in \cite{author26}.

\noindent\textbf{Evaluation metric.} We follow the standard evaluation metric (Object Keypoint Similarity--OKS\cite{author25, author26})\footnote{\url{http://cocodataset.org/\#keypoints-eval}} to measure the performance of each human pose estimation method on the COCO keypoint detection and CrowdPose datasets. Thus, for the MS COCO keypoint detection dataset, in this paper, we report their standard average precision scores with different thresholds and different object sizes and average recall scores: $AP$ (the mean of $AP$ scores at $OKS = \{0.50, 0.55,\cdots. 0.90,0.95\}$), $AP^{50}$ (AP at $OKS=0.50$), $AP^{75}$, $AP^{M}$ for medium objects, $AP^{L}$ for large objects and $AR^{50}$ (the recall scores at $OKS=0.50$). However, for the CrowdPose dataset, in addition to evaluation metrics similar to those adopted in COCO keypoint detection dataset, we also use two extra evaluation metrics: $AP^{E}$ (AP scores on relatively easier samples) and $AP^{H}$ (AP scores on harder samples).
\subsection{Implementation details}
In this paper, our proposed model belongs to the top-down human pose estimation method, which adopts the person detector provided by the SimpleBaseline \cite{author29} with 56.4\% AP for MS COCO keypoint detection dataset and follows the original paper \cite{author45} to utilize the YOLOv3 \cite{author76} as the human detector for the CrowdPose dataset. For our proposed method, the teacher model uses the SimCC method presented by the original paper \cite{author70}, which adopts HRNet-W48 as its backbone, and the first-stage student model also employs the same model, except that its backbone is HRNet-W32. In the second-stage distillation, our proposed model introduces the IGP-GCN module to refine the human pose obtained by the first-stage student model, which is trained in a progressive manner under the supervision of the same teacher model. For training the first-stage distillation, the student model follows the parameters setting in the SimCC and is trained 210 epochs. The hyper-parameters $\alpha$ and $\beta$ of the loss function (Eq.\ref{eq3}) are set to $5\times10^{-5}$ and 0.1 respectively. In addition, the student model also adopts the label smoothing in the training processing because it can make the model's generalization better. For the second-stage distillation, we train the 30 epochs, where the hyper-parameters $\delta$, $\lambda$ and $\xi$ are severally set as 0.3, 0.5 and 1, and the initial learning rate is $10^{-3}$. Our proposed model is implemented by Pytorch and all the experiments are run on a single NVIDIA A100 GPU with 40G RAM.
\subsection{Comparison with state-of-the-art HPE methods}
In this subsection, to comprehensively validate the effectiveness of our proposed model, we compare it with other state-of-the-art HPE methods on COCO \emph{val2017} and \emph{test-dev2017} and CrowdPose datasets.

\noindent\textbf{COCO \emph{val2017}}. As shown in the Table \ref{tab:val}, we compare our proposed model with some state-of-the-art and the most related HPE methods from the literature, including the classic mainstream HPE models (\emph{e.g.}, HRNet \cite{author26}, SimCC\cite{author70} and DARK\cite{author20}) and knowledge distillation-based HPE methods (\emph{e.g.}, OKDHP\cite{author46}, FPD\cite{author51}, DSPose\cite{author47} and DistilPose\cite{author48}). Seeing from the Table \ref{tab:val}, our proposed model reaches 76.2\% AP on COCO \emph{val2017} dataset, which is significantly superior to the HPE models that adopt the same backbone with it. Taking HRNet as an example, when it adopts the same backbone and input size with our HPE model, the AP score of our proposed HPE model is 1.8\% higher than the one of HRNet because our HPE model not only transfers knowledge from a large teacher model but also mines the inherent structure information of human joints. Comparing to the SimCC\cite{author70}, under the same backbone (HRNet-W32) and input size ($256\times192$), our proposed HPE model raises by 0.9\% in terms of AP score. The main reason is that the SimCC lacks the ability to model the topology of human joints and our HPE method can make full use of the inherent structure information of human joints by skeleton-aware loss and IGP-GCN module. Compared with the existing distillation-based HPE methods, our proposed HPE model balances the computational complexity and accuracy better, for example, although OKDHP (8-Stack HG) also achieves an AP score of 76.2\%, it uses a more computationally complex backbone (8-Stack HG) compared to our HPE approach (HRNet-W32 backbone).
\begin{table*}[ht]
  \caption{Comparison results on the COCO \emph{val2017} dataset. The best results are highlighted by the bold.}
  \centering
  \resizebox{1.995\columnwidth}{!}{
  \begin{tabular}{ccccccccccc}
    \toprule
     Methods & Backbone & Input size & Params(M)&GFLOPs& AP(\%) &$AP^{50}$(\%)&$AP^{75}$(\%)&$AP^{M}$(\%)&$AP^{L}$(\%)&$AR$(\%)\\
     \midrule
     \multicolumn{11}{c}{2D Heatmap-based}\\
     \midrule
     HRNet\cite{author26}&HRNet-W32&$256\times192$ &28.5&7.2 &74.4&90.5 &81.9 &70.8 &81.0&79.8\\
     HRNet\cite{author26}&HRNet-W48&$256\times192$ &63.6&14.6 &75.1&90.6 &82.2 &71.5 &81.8&80.4\\
     HRNet\cite{author26}&HRNet-W32&$384\times288$ &28.5&16.0&75.8&90.6 &82.7 &71.9 &82.8&81.0\\
     HRNet\cite{author26}&HRNet-W48&$384\times288$ &63.6&32.9 &76.3&90.8 &82.9 &72.3 &83.4&81.2\\
     DARK\cite{author20}&HRNet-32 &$256\times192$&28.5 &7.1 & 75.6&90.5 &82.1 &71.8 &82.8&80.8\\
     DARK\cite{author20}&HRNet-32 &$384\times288$&63.6 &32.9 & 76.6&90.7 &82.8 &72.7 &83.9&81.5\\
     \midrule
     \multicolumn{11}{c}{Knowledge distilation-based}\\
     \midrule
     OKDHP\cite{author46}&2-Stack HG&$256\times192$&13 &25.5 &72.8 &91.5 &79.5 &69.9 &77.1&75.6\\
     OKDHP\cite{author46}&4-Stack HG&$256\times192$&24 &47 &74.8 &92.5 &81.6 &72.1 &78.5&77.4\\
     OKDHP\cite{author46}&8-Stack HG&$256\times192$&- &- &76.2 &92.6 &83.7 &73.4 &80.2&78.8\\
     FPD\cite{author51}&HRNet-W32&$256\times192$&28.5 &7.1 &75.5 &90.6 &82.3 &71.4 &82.0&80.4\\
     SDPose-B\cite{author47}&HRNet-W32&$259\times192$&13.2 &5.2 &73.7 &89.6 &80.4 &70.3 &80.5&79.1\\
     DistilPose-L\cite{author48}&HRNet-W48-stage3&$259\times192$&21.3 &10.3 &74.4 &- &- &- &-&-\\
     \midrule
     \multicolumn{11}{c}{Transformer-based}\\
     \midrule
     TransPose-H\cite{author77}&HRNet-W48+Trans&$256\times192$&17.5 &21.8 &75.8 &- &-&- &-&80.8\\
     TFPose\cite{author80}&ResNet50+Trans &$384\times288$ &- &20.4& 72.4&- &- &- &-&-\\
     ViTPose\cite{author52}&ViTPose-B &$256\times192$&86 &17.1 & 75.8&- &- &- &-&81.1\\
     \midrule
     \multicolumn{11}{c}{SimCC-based}\\
     \midrule
     SimCC\cite{author70}&HRNet-W32 &$256\times192$&28.5 &7.8 & 75.3&- &- &- &-&80.8\\
     SimCC\cite{author70}&HRNet-W48 &$256\times192$&66.3 &14.6 & 75.9&- &- &- &-&81.2\\
     SimCC\cite{author70}&HRNet-W48 &$384\times288$&66.3 &32.9 & 76.9&90.9 &83.2 &73.2 &83.8&82\\
     Ours&HRNet-W32 &$256\times192$ &31&8.1& \textbf{76.2}&\textbf{90.8} &\textbf{82.8} &\textbf{72.0} &\textbf{82.9}&\textbf{81.1}\\
    \bottomrule
  \end{tabular}
  }
  \label{tab:val}
\end{table*}

\noindent\textbf{COCO \emph{test-dev2017}.} Table \ref{tab:test} reports the comparison results between our proposed HPE method and other state-of-the-art HPE models on COCO \emph{test-dev2017} dataset. We observe that our proposed model outperforms other HPE ones that adopt the same backbone and input size as ours, especially compared to the most related HRNet and SimCC, our model obtains the gains of 1.6\% and 0.8\% in terms of AP score, respectively. In addition, our proposed model also surpasses the OPEC-Net by the gain of 1.2\% in terms of AP score.
\begin{table*}
  \caption{Comparison results on the COCO \emph{test-dev2017} dataset.  The best results of each method are highlighted by the bold.}
  \centering
  \resizebox{1.995\columnwidth}{!}{
  \begin{tabular}{ccccccccccc}
    \toprule
     Methods & Backbone & Input size &Params(M)&GFLOPs& AP(\%) &$AP^{50}$(\%)&$AP^{75}$(\%)&$AP^{M}$(\%)&$AP^{L}$(\%)&$AR$(\%)\\
     \midrule
     \multicolumn{11}{c}{2D Heatmap-based}\\
    \midrule
     RMPE\cite{author43}&PyraNet&$320\times256$&28.1 &26.7 &72.3 &89.2 &79.1 &68.0 &78.6&-\\
     SimpleBaseline\cite{author29}&ResNet-152&$384\times288$&68.6 &35.6 &73.7 &91.9 &81.1&70.3 &80.0&79.0\\
     TransPose-H\cite{author77}&HRNet-W48+Trans&$256\times192$&17.5 &21.8 &75.0 &92.2 &82.3&71.3 &81.1&80.1\\
     HRNet\cite{author26}&HRNet-W32&$256\times192$&28.5 &7.2 &73.5 &- &-&- &-&-\\
     HRNet\cite{author26}&HRNet-W32&$384\times288$&63.6 &16.0 &74.9 &92.5 &82.8&71.3 &80.9&80.1\\
     DARK\cite{author20}&HRNet-48 &$384\times288$&63.6 &32.9 & 76.2&92.5 &83.6 &72.5 &82.4&81.1\\
     \midrule
     \multicolumn{11}{c}{Regression-based}\\
     \midrule
    SPM\cite{author26}&Hourglass &- &- &- & 66.9&88.5 &72.9 &62.6 &73.1&-\\
    TFPose\cite{author80}&ResNet50+Trans &$384\times288$&- &20.4 & 72.2&90.9 &80.1 &69.1 &78.8&-\\
    PRTR\cite{author78}&HRNet-W32+Trans &$384\times288$&41.5 &11.0 & 71.7&90.6 &79.6 &67.6 &78.4&78.8\\
    RLE\cite{author79}&HRNet-W48&$256\times192$ &28.5 &14.6 & 75.7&92.3 &82.9 &72.3 &81.3&-\\
    DistilPose-L\cite{author48}&HRNet-W48-stage3&$259\times192$&21.3 &10.3 &73.7 &91.6 &81.1 &70.2 &79.6&-\\
    OPEC-Net\cite{author68}&ResNet-152&$384\times288$&- &- &73.9 &91.9 &82.2 &- &-&-\\
     \midrule
     \multicolumn{11}{c}{SimCC-based}\\
     \midrule
     SimCC\cite{author70}&HRNet-W32 &$256\times192$&28.5 &7.8 &74.3&- &- &-&-&-\\
     SimCC\cite{author70}&HRNet-W48 &$256\times192$&66.3 &14.6 &75.4&92.4 &82.7 &71.9 &81.3&80.5\\
     Ours&HRNet-W32 &$256\times192$&31 &8.1 & \textbf{75.1}&\textbf{92.2} &\textbf{82.7} &\textbf{71.7} &\textbf{81}&\textbf{80.2}\\
    \bottomrule
  \end{tabular}
  }
  \label{tab:test}
\end{table*}

\noindent\textbf{CrowdPose.}  CrowdPose dataset has more crowded scenes than COCO keypoint dataset, which poses more challenges to the HPE models. In order to further verify the effectiveness of our proposed model in crowded scenes, Table \ref{tab:crowd} gives the quantitative comparison results on CrowdPose test set. Seeing from the Table \ref{tab:crowd}, our proposed HPE model achieves the AP score of 68.7\% on the CrowdPose dataset, which is significantly better than other comparison models. For example, on CrowdPose dataset, the AP score of our proposed HPE model is higher 2.8\% than the one of HrHRNet that adopts the more complex backbone and the larger input size, and raises by 2\% compared to the one of SimCC that uses the same with backbone and input size. The main reason is that our method mines the contextual information among human joints by utilizing the human structural constraint loss in the first-stage distillation and introducing the IGP-GCN module in the second-stage pose refinement,
which can help our model locate the invisible human joints in the challenging occlusion scenes.
\begin{table*}
  \caption{Comparison results on the CrowdPose test set. Our results are highlighted by the bold.}
  \centering
  \resizebox{1.995\columnwidth}{!}{
  \begin{tabular}{ccccccccccc}
    \toprule
     Methods & Backbone & Input size &Params(M)&GFLOPs& AP(\%) &$AP^{50}$(\%)&$AP^{75}$(\%)&$AP^{M}$(\%)&$AP^{L}$(\%)&$AR$(\%)\\
     \midrule
     Mask R-CNN\cite{author11}&ResNet101-FPN&$320\times256$ &- &- &57.2 &83.5 &60.3 &- &-&65.9\\
     AlphaPose+\cite{author45}&ResNet101&$320\times256$&- &- &66.0 &84.2 &71.5 &- &-&72.7\\
     SimpleBaseline\cite{author29}&ResNet101&$320\times256$ &- &- &60.8 &81.4 &65.7 &- &-&67.2\\
     SPPE\cite{author43}&-&-&- &- &66.0&84.2 &71.5 &75.5 &66.3&57.4\\
     HrHRNet\cite{author25}&HRNet-W48&$640\times640$&63.8 &154.3 &65.9 &86.4 &70.6 &73.3 &66.5&57.9\\
     DEKR\cite{author7}&HRNet-W48&$640\times640$&65.7 &141.5 &67.3 &86.4 &72.2 &74.6&68.1&58.7\\
     SimCC\cite{author70}&HRNet-W32 &$256\times192$&28.5&7.8 &66.7&82.1 &72.0 &74.1 &67.8&56.2\\
     Ours&HRNet-W32 &$256\times192$&31&8.1 & \textbf{68.7}&\textbf{86.7} &\textbf{73.5} &\textbf{75.5} &\textbf{68.9}&\textbf{59.3}\\
    \bottomrule
  \end{tabular}
  }
  \label{tab:crowd}
\end{table*}

\begin{figure*}
  \centering
  \includegraphics[width=17.5cm,height=6cm]{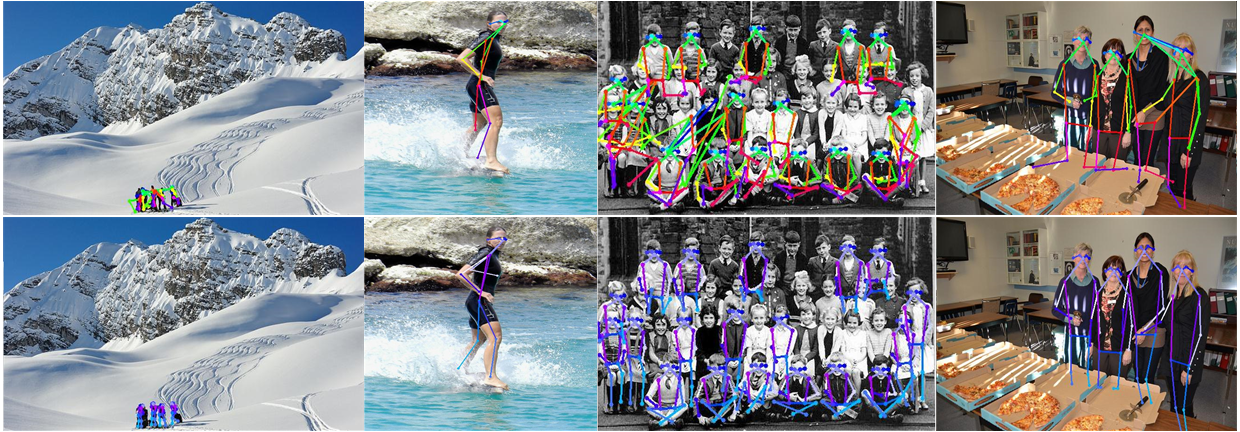}\\
  \caption{Visualizations of our proposed model and the baseline model (SimCC) on some typical example images (from left to right are respectively mountain skiing, water sports, black and white photo and indoor multi-person interaction) of COCO \emph{test-dev2017} dataset. The top row is the visualization of the baseline model (SimCC\cite{author70}), and the bottom row is the visualization of our proposed HPE model. Different person poses are painted in different colors to achieve better visualization.}\label{fig_2}
\end{figure*}

\noindent\textbf{Visualizations.} In order to more intuitively illustrate the comparison between our proposed HPE model and the baseline model (SimCC), Figure \ref{fig_2} gives their visualization results on the various scenes. Seeing from the Figure \ref{fig_2}, when dealing with small person and multi-person scenario (e.g., mountain skiing), the some human joints of the baseline model are confused and wrong connected, especially for the skiers who are close to each other. Our proposed model can clearly identify the pose of each skier, accurately connect these human joints and build the reasonable human skeleton model, which benefits from the fact that our proposed HPE model fully considers the topology of human joints. When some human joints are occluded (\emph{e.g.}, water sports and indoor multi-person interaction), the baseline model can not infer the position of the occluded joints, leading to incomplete or wrong pose connection. Our proposed HPE model can utilize the topology and contextual information of human joints to predict the position of the occluded human joints (\emph{e.g.}, the occluded left ankle in water sports), so as to maintain the complete and reasonable pose connection. For low quality image (\emph{e.g.}, black and white photo), the baseline model can not accurately identify and locate some human joints, leading to the wrong connection of human joints. However, by feature distillation and skeleton-aware distillation, we proposed PHE model can learn from the teacher model to more robust feature representation and mine the topology information between human joints, which can help our model accurately identify human joints in low-quality image and built a relatively reasonable human skeleton model.
\subsection{Ablation study}
On CrowdPose test set, we perform a number of ablation experiments to validate how each type of knowledge distillation influences the performance of our proposed HPE model. Thus, in the Table \ref{tab:ablation-1}, besides the baseline model, we also design 6 different algorithm schemes, \emph{i.e.}, for the scheme-1, it is a two-stage human pose estimation model, in which we don't adopt the knowledge distillation to transfer knowledge from the teacher model, and only introduce the image-guide progressive graph neural network to refine the coarse human pose that obtained from the baseline model. The scheme-2 is a one-stage knowledge distillation framework for human pose estimation and it only perform the feature distillation in the mid-layer of the model. The scheme-3 and scheme-4 are both the one-stage knowledge distillation framework for HPE, but different from the scheme-2, in addition to perform the feature distillation in the mid-layer of the model, they also carry out the distillation for the final output human pose. The difference between scheme-3 and scheme-4 is that the scheme-3 uses the $\ell_{1}$ loss for the pose distillation, while the scheme-4 adopts the skeleton-aware loss. Similar to the scheme-1, the scheme-5 is also a two-stage human pose estimation model, however, different from the scheme-1, in the second-stage pose refinement process, in order to refine the coarse human pose that obtained from the first-stage HPE model, the scheme-5 supervise the training of the IGP-GCN module in the progressive way by the final output pose of teacher model, that is knowledge distillation. Our scheme-6 is a two-stage knowledge distillation framework for human pose estimation, and it carries out the feature distillation of the mid-layer and skeleton-aware distillation of the final output pose in the first-stage model, and for the second stage, our method adopts the knowledge distillation strategy similar to the scheme-5 for refine the coarse pose that estimated by the first-stage model.

 Seeing from the Table \ref{tab:ablation-1}, we find that the AP score of the scheme-1 raises by 0.9\% than the one of baseline model and other two metrics are also higher than its, and the reason is that the IGP-GCN that used in the scheme-1 can utilize the image context and pose structure clues to locate the human joints more accurately. The scheme-2 is higher 0.6\% than the baseline in terms of AP score, and it turns out that the feature distillation of the mid-layer can make the student model learn more robust feature representation from a powerful teacher model, which contributes to more accurate human pose estimation. The performance of the scheme-3 is better than that of the scheme-2, but is inferior to that of the scheme-4, which suggests that the performance of the student model can be further improved when adding the pose distillation in the final output and the different pose distillation losses bring the different performance gains to the student model, \emph{e.g.}, comparing to the scheme-2, the AP score of the scheme-4 raised by 1.6\% but the one of the scheme-3 is only improved by 1.1\% because the skeleton-aware loss can make better use of the inherent structure information of human pose than the $\ell_{1}$ loss. Seeing the scheme-1 and scheme-5 in the Table \ref{tab:ablation-1}, we notice that all the evaluation metrics of the scheme-5 are superior to those of the scheme-1, which also demonstrates that the knowledge distillation can indeed improve the performance of HPE student model. Furthermore, we also observe that our proposed HPE model based on two-stage distillation is significantly better than other several HPE schemes, with an Ap score of 2.0\% higher than the baseline and 0.5\% higher than the scheme-4 (one-stage distillation model).
\begin{table*}
  \caption{Ablation study of the different schemes for our proposed HPE model on the CrowdPose test set. The best results are highlighted by the bold.}
  \centering
  \begin{tabular}{cccccccc}
    \toprule
     Methods& Features distillation & $L_{1}$ distillation & Skeleton-aware distillation & Progressive GCN pose distillation& AP(\%) &$AP^{50}$(\%)&$AP^{75}$(\%)\\
    \midrule
    Baseline & & &  &  &66.7 &82.1&72.0\\
    scheme-1 & & &  & &67.6 &85.2&72.7\\
    scheme-2 &\checkmark & &  &  &67.3 &84.1&72.3\\
    scheme-3 &\checkmark &\checkmark &  &  &67.7 &84.8&72.6\\
    scheme-4 &\checkmark & &\checkmark  &  &68.2 &86.3&73.3\\
    scheme-5 &  &  &  &\checkmark &67.9 &85.8&73.0\\
    scheme-6\\(ours) &\checkmark  &  &\checkmark  & \checkmark &\textbf{68.7} &\textbf{86.7}&\textbf{73.5}\\
    \bottomrule
  \end{tabular}
  \label{tab:ablation-1}
\end{table*}

\section{Conclusion}\label{sec:con}
 In this paper, we propose a novel coarse-to-fine two-stage knowledge distillation framework for human pose estimation. Our model not only transfers high-level semantic knowledge from a powerful teacher model but also makes full use of the structure information among human joints so as to accurately locate the human joints. The extensive experiments on COCO keypoint and CrowdPose datasets demonstrate that our proposed HPE method outperforms lots of the existing state-of-the-art HPE methods,  especially for the more complex CrowdPose dataset, the performance improvement of our model is more significant.


%

%

\section*{Acknowledgment}
This work is supported by the National Natural Science Foundation of China (No.61602288), Fundamental Research Program of Shanxi Province (No.20210302123443, 202203021221002), and National Key Research and Development Program of China
(No.2020AAA0106100). The authors also would like to thank the anonymous reviewers for their valuable suggestions.

\ifCLASSOPTIONcaptionsoff
  \newpage
\fi



%
{\small
\bibliographystyle{IEEEtran}
\bibliography{referenceBib}
}
%
%

%

\begin{IEEEbiography}[{\includegraphics[width=1in,height=1.25in,clip,keepaspectratio]{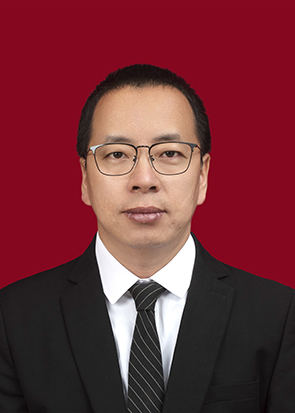}}]{Zhangjian Ji}
received the B.S. degree from Wuhan University, China, in 2007, the M.E. degrees from the Institute of Geodesy and Geophysics, Chinese
Academy of Sciences (CAS), China, in 2010, and the Ph.D. degree in the University of Chinese Academy of Sciences, China, in 2015.

He is currently an Associate Professor at the School of Computer and Information Technology, Shanxi University, Taiyuan, China. His research interests include computer vision, pattern recognition, machine learning and human-computer interaction.
\end{IEEEbiography}
\vspace{-5mm}
\begin{IEEEbiography}[{\includegraphics[width=1in,height=1.25in,clip,keepaspectratio]{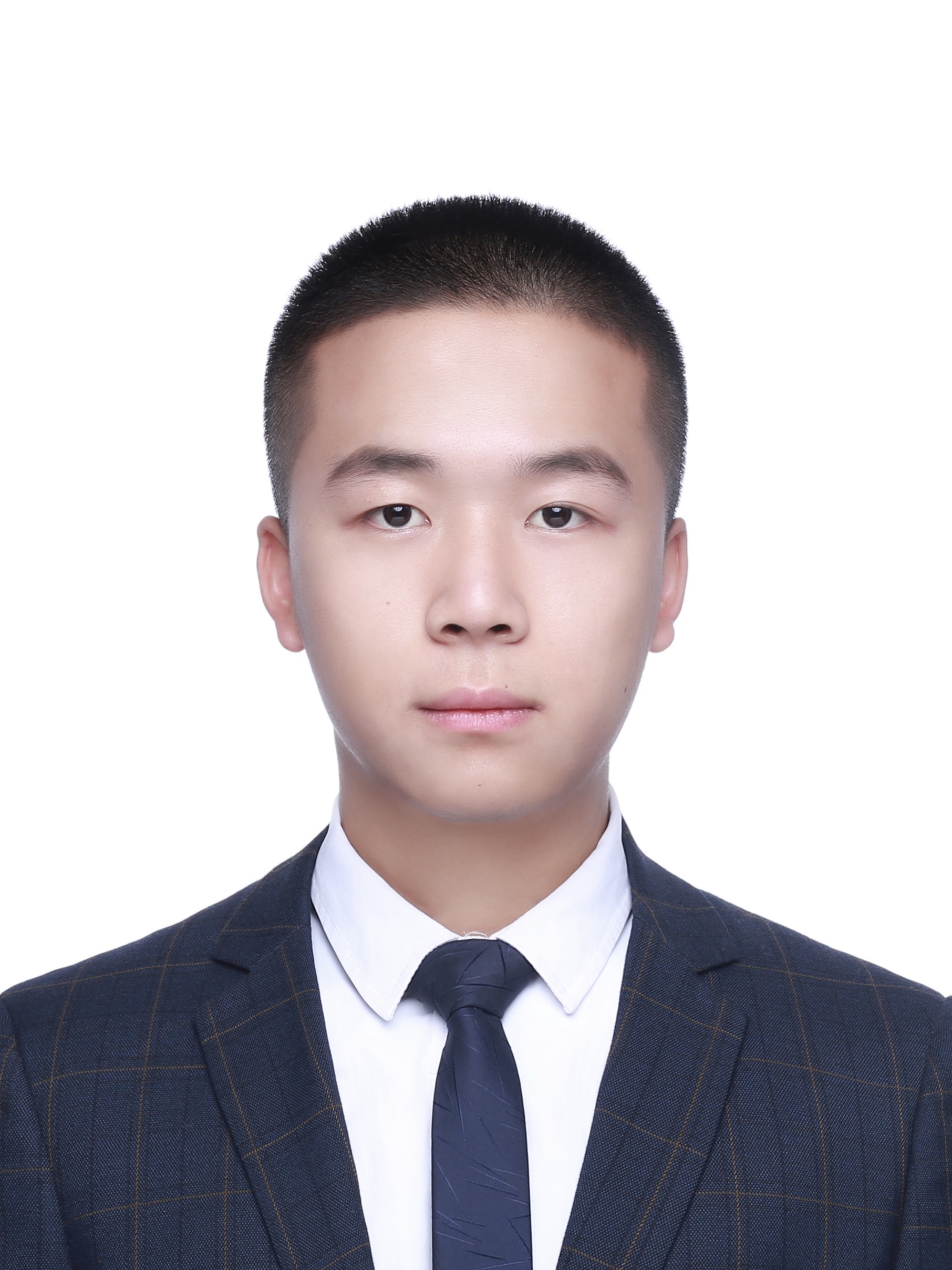}}]{Wenjin Zhang}
received the B.S. degree in engineering from Shanxi Agricultural University, China, in 2022. received the M.E. degree in engineering from Shanxi University, China, in 2025.

His research interests include computer vision, human-computer interaction, etc.
\end{IEEEbiography}
\begin{IEEEbiography}[{\includegraphics[width=1in,height=1.25in,clip,keepaspectratio]{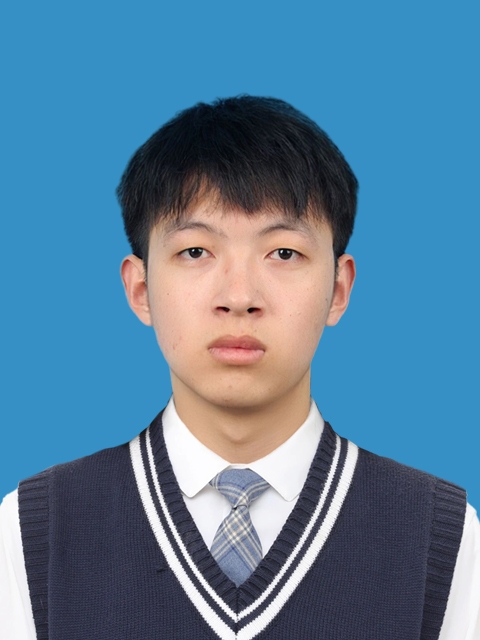}}]{ Shaotong Qiao}
received the B.S. degree in engineering from Shandong Jianzhu University, China, in 2020. He is currently pursuing the M.E. degrees in Computer technologies from Shanxi university, China.

His research interests include computer vision, person re-identification, etc.
\end{IEEEbiography}
\begin{IEEEbiography}[{\includegraphics[width=1in,height=1.25in,clip,keepaspectratio]{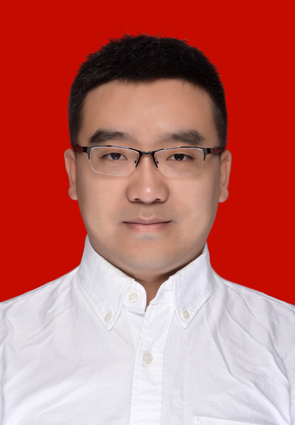}}]{Kai Feng}
received the PhD degree from Shanxi University, China, in 2014. He is currently an Associate Professor at the School of Computer and Information Technology, Shanxi University, China. His research interests include combinatorial optimization and interconnection network analysis.
\end{IEEEbiography}
\vspace{-13cm}
%
\begin{IEEEbiography}[{\includegraphics[width=1in,height=1.25in,clip,keepaspectratio]{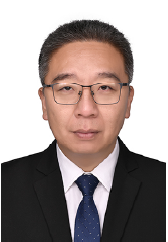}}]{Yuhua Qian}
is a professor and Ph.D. supervisor of Key Laboratory of Computational Intelligence and Chinese Information Processing of Ministry
of Education, China. He received the M.S. degree and the PhD degree in Computers with applications at Shanxi University in 2005 and 2011, respectively. He is best known for multigranulation rough sets in learning from categorical data and granular computing. He is actively pursuing research in pattern recognition, feature selection, rough set theory, granular computing and artificial intelligence. He has published more than 200 articles on these topics in international top journals or conferences including AI, IEEE TPAMI, JMLR, IEEE TKDE, NIPS, AAAI, IJCAI, etc. He served on the Editorial Board of the International Journal of Knowledge-Based Organizations and Artificial Intelligence Research. He has served as the Program Chair or Special Issue Chair of the Conference on Rough Sets and Knowledge Technology, the Joint Rough Set Symposium, and the International Conference on Intelligent Computing, etc., and also PC Members of many machine learning, data mining conferences.
\end{IEEEbiography}
%
%




\end{document}